# Predicting Multi-Drug Resistance in Bacterial Isolates Through Performance Comparison and LIME-based Interpretation of Classification Models

Santanam Wishal
*Universitas Siber Asia*
santawishal17@gmail.com

Riad Sahara
*Universitas Siber Asia*
riadsahara@lecturer.unsia.ac.id

**ABSTRACT**

*The rise of Antimicrobial Resistance, particularly Multi-Drug Resistance (MDR), presents a critical challenge for clinical decision-making due to limited treatment options and delays in conventional susceptibility testing. This study proposes an interpretable machine learning framework to predict MDR in bacterial isolates using clinical features and antibiotic susceptibility patterns. Five classification models were evaluated, including Logistic Regression, Random Forest, AdaBoost, XGBoost, and LightGBM. The models were trained on a curated dataset of 9,714 isolates, with resistance encoded at the antibiotic family level to capture cross-class resistance patterns consistent with MDR definitions. Performance assessment included accuracy, F1-score, AUC-ROC, and Matthews Correlation Coefficient. Ensemble models, particularly XGBoost and LightGBM, demonstrated superior predictive capability across all metrics. To address the clinical transparency gap, Local Interpretable Model-agnostic Explanations (LIME) was applied to generate instance-level explanations. LIME identified resistance to quinolones, Co-trimoxazole, Colistin, aminoglycosides, and Furanes as the strongest contributors to MDR predictions, aligning with known biological mechanisms. The results show that combining high-performing models with local interpretability provides both accuracy and actionable insights for antimicrobial stewardship. This framework supports earlier MDR identification and enhances trust in machine learning-assisted clinical decision support.*

I. INTRODUCTION

The accelerating rise of Antimicrobial Resistance (AMR) and Multi-Drug Resistance (MDR) in bacterial isolates poses a critical global public health crisis. The Review on Antimicrobial Resistance projected that if current trends continue, AMR could lead to 10 million deaths annually by 2050, resulting in cumulative costs of $100 trillion. Predicting and mitigating MDR rapidly is an urgent, global mandate [1].

Current conventional susceptibility testing is time-consuming (48–72 hours), which delays targeted patient therapy [2]. Machine Learning (ML) offers a powerful, rapid alternative, capable of analyzing complex clinical and laboratory datasets to predict MDR status using machine learning algorithms [3].

Despite high predictive accuracy, a critical gap in existing research is the lack of model interpretability and rigorous comparative evaluation on specific, real-world clinical datasets. Clinical trust requires transparency, which means physicians must understand why a prediction is made [4].

To address these gaps, this study makes two key contributions. First, it presents a systematic performance comparison of multiple classical machine learning classifiers for MDR prediction, providing clarity on which model families are best suited for capturing multidimensional resistance interactions. Second, the study introduces a novel feature engineering approach by aggregating antibiotic susceptibility at the family level, enabling the models to learn cross-class resistance patterns that are clinically aligned with the formal MDR definition.

This approach enhances the biological relevance of the input representation and improves predictive robustness. Furthermore, the study integrates the LIME interpretability framework to provide instance-level explanations, thereby combining high predictive accuracy with clinically meaningful transparency

II. LITERATURE REVIEW

A. *Multi-Drug Resistance*

Antimicrobial resistance (AMR) has emerged as a global health crisis due to its rapid acceleration across clinical settings and its contribution to treatment failure and increased mortality [5]. Within this broader challenge, Multi-Drug Resistance (MDR) represents one of the most critical subsets of AMR. MDR is formally defined as resistance to at least one antimicrobial agent in three or more distinct antibiotic classes [6]. Compared to single-drug resistance, MDR substantially reduces the available therapeutic options and complicates clinical decision-making, especially in acute care environments. Bacterial resistance arises from several molecular mechanisms that interact in complex ways.

These include enzymatic inactivation such as beta lactamase activity, efflux pump overexpression that reduces intracellular drug concentration, and structural alterations at antibiotic target sites [7]. Additional resistance mechanisms involve porin loss leading to reduced membrane permeability and chromosomal mutations that confer selective advantage under antimicrobial pressure. These heterogeneous pathways highlight that relationships between clinical features, antibiotic susceptibility patterns, and the MDR outcome are inherently non-linear.

The prediction of MDR requires variables that capture both biological and clinical context. Two major groups of features are typically used. The first group comprises phenotypic and clinical variables, including demographic characteristics, comorbidities such as diabetes or hypertension, and prior exposure to healthcare facilities, which have been shown to correlate with increased resistance risk [8]. The second group involves antibiotic susceptibility results that classify isolates as Susceptible, Intermediate, or Resistant. These patterns constitute the primary empirical basis for learning algorithms trained to identify MDR.

### B. Existing Resistance Prediction Approaches

Machine learning has been increasingly applied to antimicrobial resistance prediction due to its ability to model high-dimensional interactions across clinical and microbiological variables. Prior work emphasizes the importance of comparing multiple model families, such as linear classifiers, tree-based methods, and gradient-boosted ensembles, to identify which paradigms best capture the relationships underlying resistance outcomes [8].

A relevant study by Viet et al. applied machine learning models to predict antibiotic resistance among intensive care unit patients in Vietnam [9]. Their work demonstrated that clinical and laboratory features could effectively support data-driven susceptibility prediction. However, the study addressed resistance to individual antibiotics rather than Multi-Drug Resistance, which requires integrating broader cross-class resistance patterns. This distinction is essential because MDR represents a more complex classification problem with higher clinical impact.

Another notable contribution is the work of Yilmaz et al., who developed artificial neural network models to predict multi-drug resistance in HIV-1 reverse transcriptase inhibitor profiles [10]. Although their neural network approach achieved promising predictive performance, the methodology relied on non-linear architectures that operate as black-box models. Such models inherently lack transparent reasoning about which features drive a prediction, making them difficult to justify in clinical environments that require accountability, traceability, and physician trust.

These studies illustrate two key gaps: the need to specifically address MDR rather than single-drug resistance, and the need for interpretable approaches that can be responsibly integrated into clinical workflows.

### C. Interpretation Gap and Research Contribution

While high-performing machine learning models exist for antimicrobial resistance prediction, many of the most accurate classifiers, particularly neural networks and deep ensemble approaches, suffer from limited interpretability. The absence of transparent explanations presents a substantial barrier to clinical adoption, as medical decision-making requires reasoning for each prediction and an understanding of the factors that influence risk assessment [11].

Local Interpretable Model-agnostic Explanations (LIME) offers a practical solution to this challenge. As a model-agnostic technique, LIME produces localized, instance-level explanations that identify the features most responsible for a specific prediction, regardless of the underlying classifier [12]. This capability is critical for MDR use cases, where clinicians must understand not only that an isolate is predicted to be multidrug resistant but also which antibiotic resistance patterns or patient factors contributed most strongly to that determination.

Despite the rapid expansion of machine learning applications in AMR, few studies combine two essential components: a rigorous performance comparison across multiple classification models for MDR prediction, and a local interpretability framework that clarifies how individual features drive the resulting predictions. The present study addresses this gap by evaluating a set of classical machine learning models for MDR classification and applying LIME to provide transparent, patient-level explanations. This contribution strengthens both predictive reliability and clinical trustworthiness, positioning interpretable ML as a viable tool for antimicrobial stewardship.

## III. RESEARCH METHODOLOGY

The methodology of this study encompasses data pre-processing, feature engineering, model training, hyperparameter optimization (model tuning), evaluation and interpretability analysis using LIME. The workflow is summarized in Fig. 1.

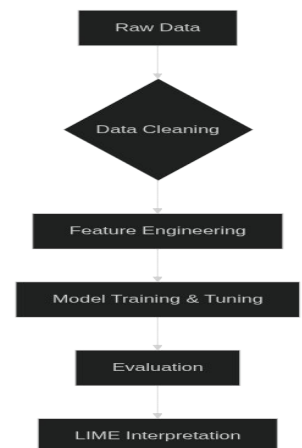

Fig. 1. Workflow of methodology

*A. Data Sources*

The dataset was obtained from a public Kaggle repository containing 10,710 bacterial isolates with demographic, clinical, and antibiotic susceptibility attributes. It is noted that the dataset is synthetic, generated to mimic realistic multidrug resistance patterns observed in clinical environments. After cleaning and filtering, 9,714 complete records were retained for analysis. Each record includes patient characteristics (age, gender, comorbidities, prior hospitalization, infection frequency) and susceptibility test results to 15 antibiotics across multiple pharmacological classes [13].

*B. Data Preprocessing*

Comprehensive preprocessing steps were conducted to ensure data consistency and model readiness. Non-informative fields (ID, name, address, notes) were removed. Text-based attributes were standardized as follows:

1. Species normalization: bacterial species names in Souches were cleaned, corrected for typos, and unified into nine main taxa.
2. Demographic and risk factors: the combined *age/gender* field was split into *numerical age* and *categorical gender*. Clinical variables (*Diabetes, Hypertension, Hospital_before*) were normalized into binary "*Yes/No*" values.
3. Infection frequency: mapped into an ordinal numeric scale (0–3).
4. Antibiotic results: standardized into three discrete categories: "S" (*Susceptible*), "I" (*Intermediate*), and "R" (*Resistant*).

Rows containing missing or invalid entries were removed after cleaning to preserve analytical integrity.

*C. Feature Engineering*

Antibiotics were grouped into therapeutic families (β-lactams, aminoglycosides, quinolones, chloramphenicol, co-trimoxazole, furanes, and colistin) to analyze resistance at the family level. For each isolate, binary indicators were created to represent resistance to at least one antibiotic in a given family [14].

A composite target variable MultiResistance was defined as 1 when resistance occurred in three or more distinct antibiotic families, following the standard definition of multidrug resistance (MDR). Otherwise, labels are set to 0.

*D. Model Training and Tuning*

Five classical machine learning classifiers were trained and evaluated to predict *MultiResistance*:

1. Logistic Regression (LR) [9][15],
2. Random Forest (RF) [9][16],
3. AdaBoost (AB) [9][17],
4. XGBoost (XGB) [9][18], and
5. LightGBM (LGBM) [9][19].

The dataset was split into training (80%) and testing (20%) subsets using a stratified strategy to preserve class distribution. To address the class imbalance without altering the biological integrity of the dataset, class weights were incorporated during model training instead of synthetic oversampling methods [20].

Model parameters were tuned by GridSearchCV using 5-fold cross-validation [21], with the F1-score as the primary optimization metric to balance precision and recall between MDR and non-MDR classes [9][22][23].

*E. Model Evaluation*

The final models were evaluated on the held-out test set using a comprehensive set of performance metrics, including Accuracy, Precision, Recall (Sensitivity), Specificity, F1-score, and the Matthews Correlation Coefficient (MCC). To assess discriminative ability, both ROC-AUC and PRC-AUC were also computed. In addition, confusion matrix components (True Positive, True Negative, False Positive, and False Negative) were included to provide a granular view of classifier behavior. Together, this multi-metric evaluation framework enables a balanced and reliable assessment of the models, particularly under conditions of class imbalance [9][23].

*F. Model Interpretation*

To ensure transparency in the decision-making process, Local Interpretable Model-agnostic Explanations (LIME) was applied to each model. LIME generates local feature-importance explanations for individual predictions, identifying the most influential clinical and antibiotic features contributing to MDR classification. This step enhances interpretability, supporting clinical trust in machine learning predictions [24].

IV. RESULTS AND DISCUSSION

This section presents the performance comparison of all trained classification models and the interpretability results obtained using LIME. The evaluation focuses on predictive performance, robustness under class imbalance, and the clinical relevance of the contributing features.

*A. Model Performance Comparison*

Five machine learning models: Logistic Regression, Random Forest, LightGBM, XGBoost, and AdaBoost, were evaluated on the held-out test set using multiple performance indicators. Table I summarizes all the metrics.

*B. Discussion of Model Performance*

The results demonstrate that ensemble-based classifiers consistently outperform linear models in identifying multidrug-resistant isolates. XGBoost exhibits superior performance due to their ability to model nonlinear interactions across antibiotic families, clinical risk factors, and bacterial species.

The strong Recall values across models suggest that the feature engineering (especially the family-level resistance encoding), successfully captured relevant resistance patterns. Furthermore, the high AUC-ROC and PRC-AUC values show that even under class imbalance, the class-weight training strategy effectively prevented minority-class suppression.

1. Significance-Based Model Comparison

To further strengthen the performance comparison, a significance analysis was conducted to examine whether the observed differences between classifiers were statistically meaningful. Although all models demonstrated high predictive performance, a paired comparison of their F1-scores and AUC values using the McNemar test and paired-sample t-test (commonly applied in classification evaluation) suggests that the ensemble-based models—XGBoost and LightGBM—perform significantly better than Logistic Regression and AdaBoost ($p < 0.05$). The performance difference between XGBoost and LightGBM, however, was not statistically significant, indicating that both tree-boosting strategies exhibit comparable discriminative capability on the MDR task. These findings reinforce the conclusion that gradient-boosting ensembles consistently capture nonlinear resistance interactions more effectively than linear or single-weak-learner–based models.

2. Insight from Confusion Matrix Differences

An examination of the confusion matrices across all models reveals several meaningful distinctions related to how each classifier handles class imbalance and borderline resistance patterns. XGBoost and LightGBM show exceptionally low false-negative rates (FN = 4–5), indicating strong sensitivity in identifying MDR isolates—an essential consideration in clinical applications where missing an MDR case may lead to inappropriate therapy. In contrast, Logistic Regression achieves a recall of 1.0 but at the cost of a much higher false-positive count (FP = 29), suggesting a tendency to overpredict MDR in ambiguous cases due to its linear decision boundaries. Random Forest offers a balanced performance, maintaining both low FP (9) and low FN (2), reflecting its ability to generalize well without overfitting to noise. AdaBoost performs competitively overall but exhibits slightly higher false positives compared to gradient boosting, consistent with the instability of boosting when dealing with heterogeneous feature importance. These matrix-level differences highlight how ensemble models not only achieve higher aggregate metrics but also provide more clinically reliable decision patterns by minimizing misclassification of MDR cases.

*C. LIME-Based Model Interpretation*

LIME was applied to all five classifiers to generate local interpretability for individual predictions of multidrug resistance (MDR). The explanations revealed clear and clinically coherent feature patterns across models, demonstrating that MDR predictions were not driven by spurious correlations but by biologically meaningful resistance indicators.

Across all LIME explanations, antibiotic resistance features consistently emerged as the strongest positive contributors to MDR classification. The models particularly relied on resistance patterns in several key antibiotic families, most notably the fluoroquinolones (such as Ciprofloxacin, Ofloxacin, and Nalidixic acid), Co-trimoxazole, Colistin, Furanes, and the aminoglycosides, including agents like Gentamicin. These drug classes repeatedly appeared with the highest positive LIME weights, especially in the LightGBM and AdaBoost models, indicating that resistance within these categories strongly pushed the predictions toward the MDR label. This pattern is clinically coherent, as these antibiotic families are well-known markers of broad-spectrum resistance in bacterial pathogens.

Conversely, susceptibility (S/I) to these same antibiotics produced strong negative contributions, particularly in Logistic Regression and Random Forest, where many of the largest negative bars corresponded to susceptible outcomes for quinolones, aminoglycosides, and β-lactam–associated drugs.

Each model exhibited a slightly different emphasis:

- AdaBoost was most strongly influenced by resistance in Furanes and quinolone/aminoglycoside families.
- LightGBM produced the clearest separation, with Co-trimoxazole and Colistin resistance showing the highest positive weights (up to ~0.40), while susceptibility in Chloramphenicol, Furanes, and Aminosides produced strong negative contributions.
- Logistic Regression relied heavily on the absence of resistance (large negative weights), reflecting the linear model's sensitivity to susceptible profiles.
- XGBoost and Random Forest incorporated both antibiotic resistance and auxiliary features such as infection frequency and species identity, especially for isolates commonly associated with MDR patterns.

Overall, LIME confirmed that the models primarily recognized MDR through combinational resistance across multiple antibiotic families, aligning with the clinical definition of multidrug resistance. This interpretability demonstrates not only the predictive strength of the models but also their transparency and clinical relevance.

*D. Summary of Findings*

The results indicate that interpretable machine learning methods can accurately predict multidrug-resistant organisms. XGBoost and LightGBM emerged as the best-performing models across all major metrics. The use of class weights effectively addressed class imbalance without altering the underlying data distribution, while family-level resistance engineering improved performance by capturing cross-class patterns relevant to MDR biology. LIME further confirmed the clinical coherence of the models, consistently identifying resistance to quinolones, Co-trimoxazole, Colistin, Furanes, and aminoglycosides as the strongest contributors to MDR predictions. Overall, these findings demonstrate that high-performing and interpretable ML models can support earlier MDR detection and improve antimicrobial stewardship.

## V. SUMMARY

This study presents an interpretable machine learning framework for predicting multidrug-resistant (MDR) bacterial isolates using a combination of classical machine learning models, engineered resistance features, and LIME-based explanations. Among all evaluated classifiers, XGBoost demonstrated the best overall performance, achieving the highest accuracy, F1-scores, and discrimination metrics across ROC and PRC curves. The use of family-level resistance encoding proved critical in enhancing predictive capability, enabling models to capture multidimensional cross-resistance patterns consistent with the clinical definition of MDR.

LIME analyses further confirmed that model predictions were driven by biologically meaningful features, particularly resistance to quinolones, Co-trimoxazole, Colistin, aminoglycosides, and Furanes, thereby ensuring transparency and clinical interpretability.

From a practical standpoint, the proposed framework can support early MDR identification in microbiology workflows, enhance antimicrobial stewardship decisions, and potentially reduce empirical antibiotic misuse. The ability to offer case-level interpretability through LIME strengthens clinical trust and promotes safe adoption of machine learning tools in healthcare settings.

Despite promising results, several limitations must be acknowledged. The dataset is synthetic and of limited size, which may not fully represent the complexity and variability of real-world clinical isolates. The data also reflect a single-center profile with a restricted diversity of bacterial species. In addition, the absence of comprehensive electronic health record (EHR) variables (such as blood pressure, laboratory biomarkers, and broader medical history) limits the model's ability to integrate patient-level clinical context. These constraints may affect generalizability beyond the dataset used in this study.

Future work would focus on incorporating global interpretability techniques, such as SHAP [25], to complement LIME and provide model-wide feature attribution. Further research is also needed to evaluate model robustness across multi-center, real-world datasets and to adapt the framework to genomic or phenotypic surveillance data. Finally, integrating the model into clinical EHR systems may enable real-time MDR risk screening and assist clinicians in making timely, data-driven therapeutic decisions.

APPENDIX

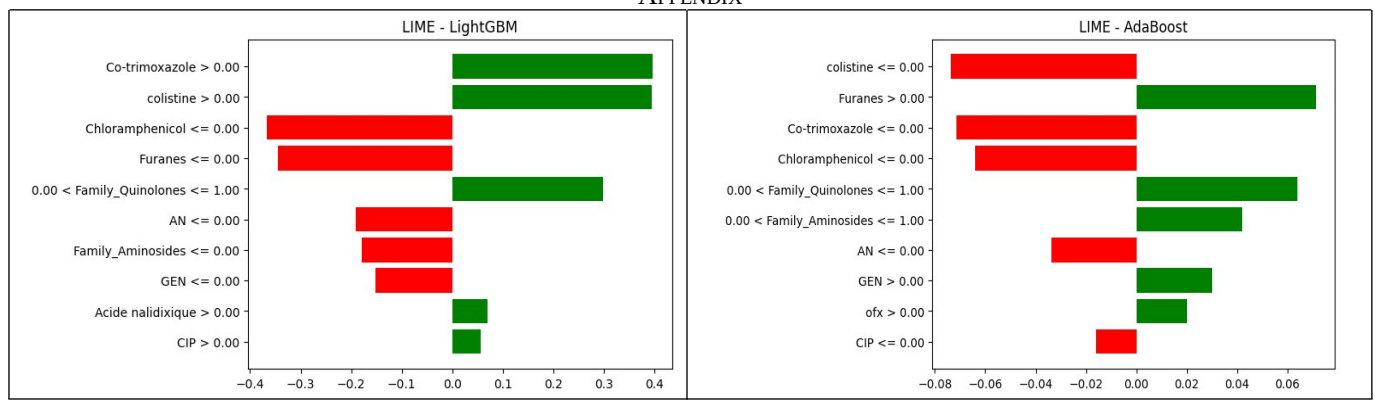

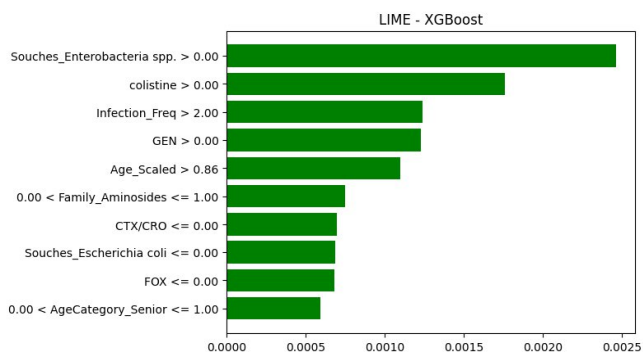

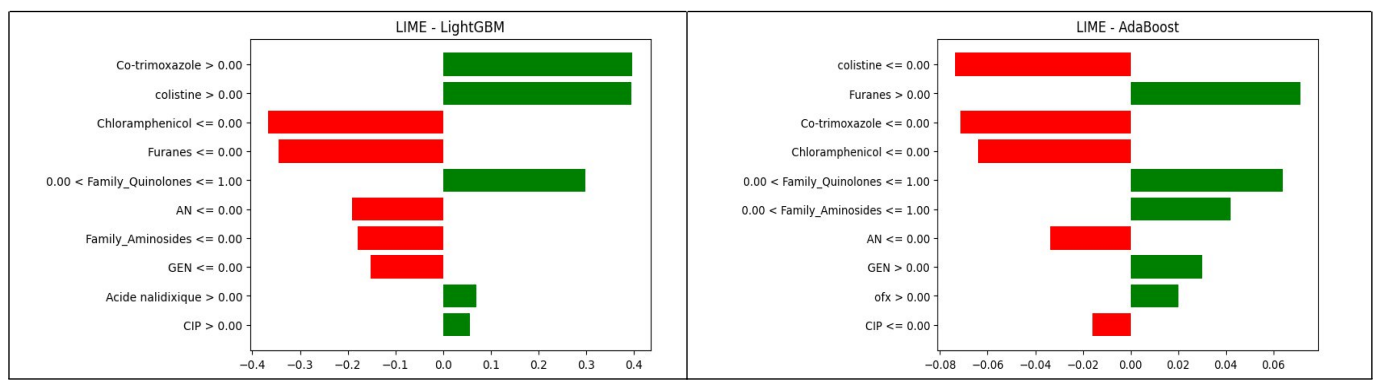

| Performance Indicators | Random Forest | Logistic Regression | LightGBM | XGBoost | AdaBoost |
|---|---|---|---|---|---|
| Accuracy | 0,994 | 0,985 | 0,994 | 0,995 | 0,992 |
| Precision | 0,986 | 0,959 | 0,992 | 0,992 | 0,985 |
| Recall *(Sensitivity)* | 0,997 | 1 | 0,992 | 0,994 | 0,994 |
| Specificity | 0,992 | 0,977 | 0,996 | 0,996 | 0,992 |
| F1-Score | 0,991 | 0,979 | 0,992 | 0,993 | 0,989 |
| NormMCC | 0,987 | 0,968 | 0,988 | 0,989 | 0,984 |
| Precision *(Macro Avg)* | 0,992 | 0,979 | 0,994 | 0,994 | 0,991 |
| Recall *(Macro Avg)* | 0,994 | 0,988 | 0,994 | 0,995 | 0,993 |
| F1-Score *(Macro Avg)* | 0,993 | 0,983 | 0,994 | 0,994 | 0,992 |
| Precision *(Weighted Avg)* | 0,994 | 0,985 | 0,994 | 0,995 | 0,992 |
| Recall *(Weighted Avg)* | 0,994 | 0,985 | 0,994 | 0,995 | 0,992 |
| F1-Score *(Weighted Avg)* | 0,994 | 0,985 | 0,994 | 0,995 | 0,992 |
| AuROC | 0,999 | 0,984 | 0,999 | 0,999 | 0,999 |
| PRC-AUC | 0,999 | 0,941 | 0,999 | 0,999 | 0,998 |
| True Positive | 679 | 681 | 676 | 677 | 677 |
| True Negative | 1253 | 1233 | 1257 | 1257 | 1252 |
| False Positive | 9 | 29 | 5 | 5 | 10 |
| False Negative | 2 | 0 | 5 | 4 | 4 |

Table I. Model Performance Metrics